\pdfoutput=1

\documentclass[11pt]{article}
\usepackage[preprint]{acl}

\usepackage{times}
\usepackage{latexsym}

\usepackage[T1]{fontenc}

\usepackage[utf8]{inputenc}

\usepackage{microtype}

\usepackage{inconsolata}

\usepackage{graphicx}

\usepackage{amsmath}
\usepackage{booktabs}
\usepackage{multirow}
\usepackage{enumitem}
\usepackage{algorithm}
\usepackage{algpseudocode}
\usepackage{colortbl}

\title{OkraLong: A Flexible Retrieval-Augmented Framework for Long-Text Query Processing}

\author{Yulong Hui$^{1}$,\ Yihao Liu$^{1}$,\ Yao Lu$^{2}$,\ Huanchen Zhang$^{1}$\\
$^1$Tsinghua University; $^2$National University of Singapore\\
\\
}

\begin{document}
\maketitle
\begin{abstract}
Large Language Models (LLMs) encounter challenges in efficiently processing long-text queries, as seen in applications like enterprise document analysis and financial report comprehension. While conventional solutions employ long-context processing or Retrieval-Augmented Generation (RAG), they suffer from prohibitive input expenses or incomplete information. Recent advancements adopt context compression and dynamic retrieval loops, but still sacrifice critical details or incur iterative costs.
To address these limitations, we propose OkraLong, a novel framework that flexibly optimizes the entire processing workflow. Unlike prior static or coarse-grained adaptive strategies, OkraLong adopts fine-grained orchestration through three synergistic components: analyzer, organizer and executor. 
The analyzer characterizes the task states, which guide the organizer in dynamically scheduling the workflow. The executor carries out the execution and generates the final answer. 
Experimental results demonstrate that OkraLong not only enhances answer accuracy but also achieves cost-effectiveness across a variety of datasets.

\end{abstract}


\section{Introduction}

Large Language Models (LLMs) have been extensively utilized to handle external knowledge and unseen data,  which is a common scenario in real-world applications such as enterprise search and data analysis (\citealp{gao2023rag-survey}; \citealp{uda}).
A critical challenge in these domains, however, lies in querying and comprehending long-form text \citep{bai-etal-2024-longbench}. For example, a company may need to query its proprietary technique documents; a financial expert may need to extract insights from the latest corporate reports; and a research group may need to assimilate cutting-edge academic papers to guide their innovations. 

To tackle long-text querying, two prevalent methodologies are typically utilized: long-context (LC) and Retrieval-Augmented Generation (RAG) (\citealp{li2024long}; \citealp{xu2024long_context}). The LC approach leverages the LLM's inherent ability to process extensive texts by inputting entire content, enabling responses grounded in global contextual awareness (\citealp{xu2024long_context}; \citealp{fei-etal-2024-extend-lc}). In contrast, RAG employs a lightweight retriever to first identify query-relevant text segments, which are then analyzed by the LLM to get the answer (\citealp{rag}; \citealp{adarag}; \citealp{selfrag}).

\begin{figure}[t]
  \includegraphics[width=\columnwidth]{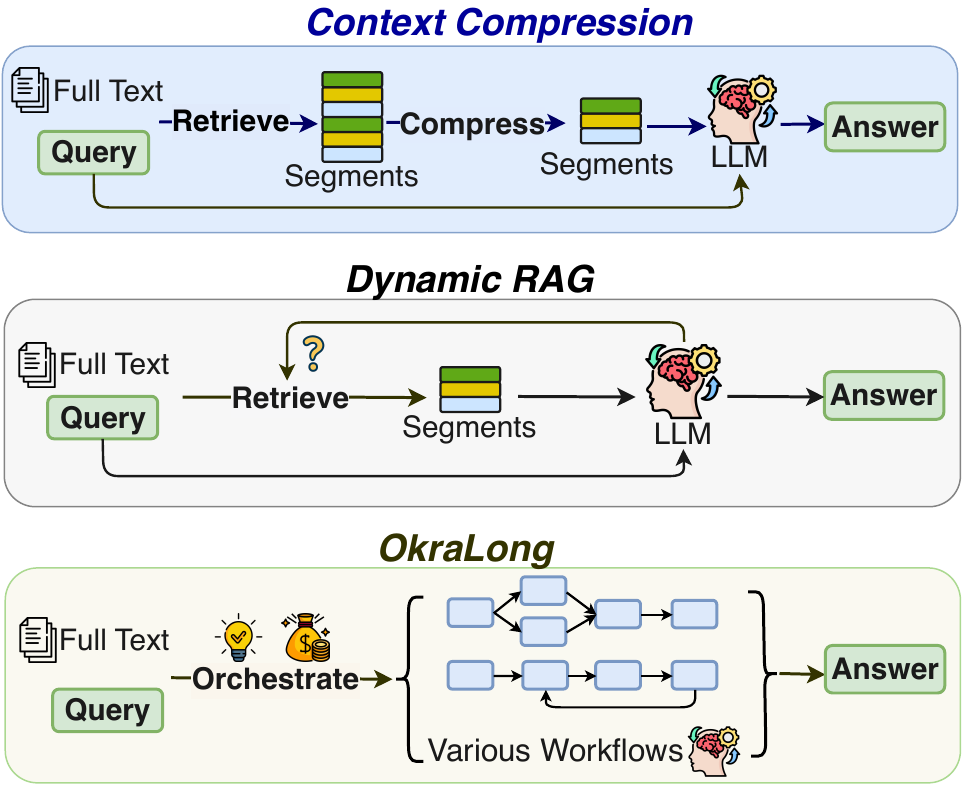}
  \caption{Comparison of OkraLong with two prevalent advanced paradigms for processing long-text queries.}
  \label{fig:intro}
\end{figure}

However, when deployed in practical settings, these strategies encounter significant limitations in cost-effectiveness and accuracy. First, given that the cost of LLMs escalate with data volume, employing LC with voluminous text may prove prohibitively expensive (e.g., a financial reports may span hundreds of pages) (\citealp{li2024self-router}; \citealp{longllmlingua}). While RAG reduces input length by filtering irrelevant text, the context content still incur moderate costs and risk omitting critical information \citep{gao2023rag-survey}. Futhermore,  real-world queries vary widely, from simple fact extraction to multi-step reasoning. Rigid approaches like static retrieval struggle to adapt to this diversity, leading to information loss and inaccuracies (\citealp{shao-etal-2023-iterative-rag}; \citealp{zhuang-etal-2024-efficientrag}).

Recent efforts to mitigate these limitations primarily focus on two paradigms: context compression and dynamic RAGs. As shown in Figure~\ref{fig:intro}, compression-based approaches typically operate on extensive text segments, removing non-informative tokens or iteratively summarize the content using generative models (\citealp{longllmlingua}; \citealp{llmlingua}; \citealp{compact}). However, these methods risk losing critical specific details and incur latency overhead due to heavy reliance on local models \citep{exit}. Dynamic RAG approaches employ iterative retrieval-generation cycles to adaptively make retrieval decisions (\citealp{selfrag}; \citealp{flare}; \citealp{dragin}). However, iterative workflow requires frequent LLM calls, escalating financial costs, and the existing adaptive mechanisms remain coarse-grained, failing to optimize the performance effectively in varied scenarios.

\begin{figure*}[t]
\centering
  \includegraphics[width=0.98\linewidth]{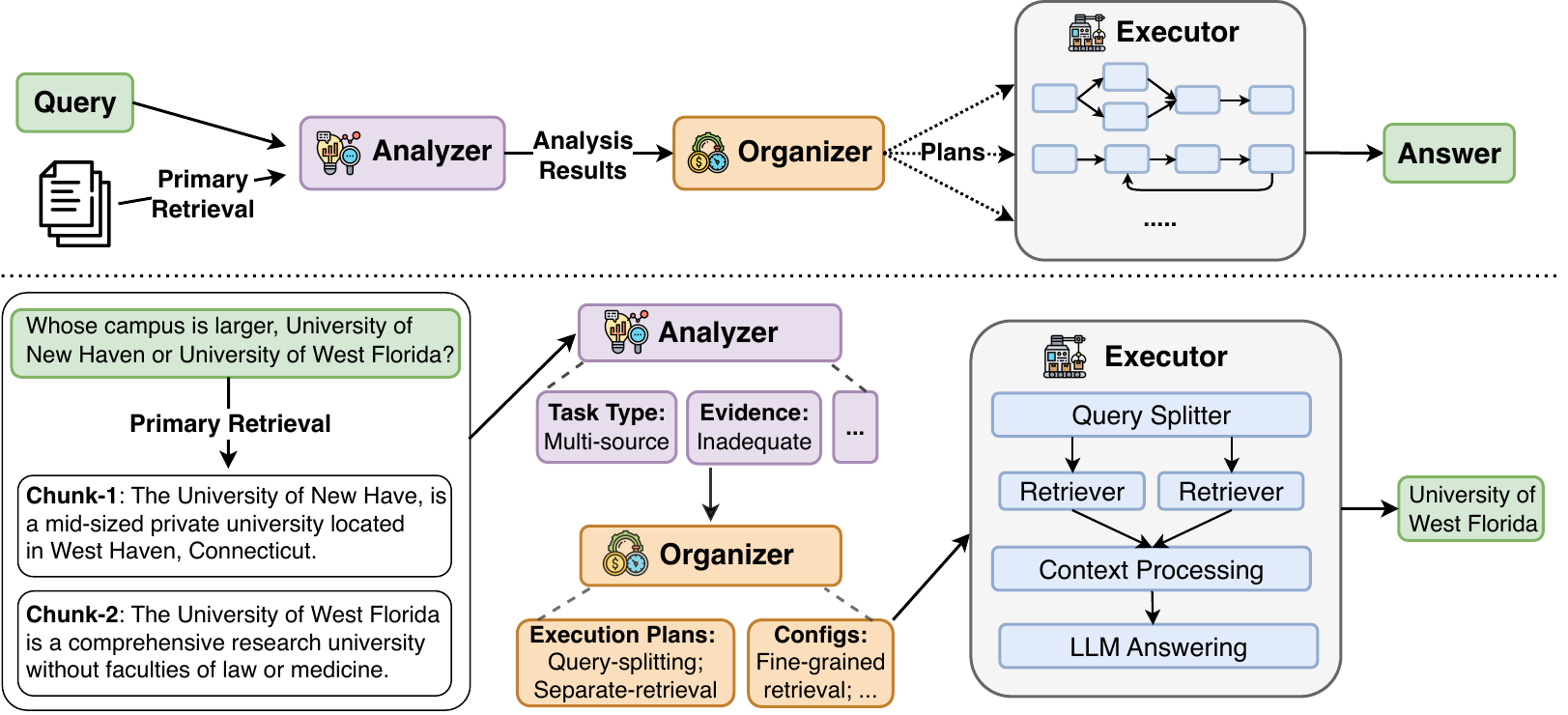}
  \caption {Architecture of OkraLong illustrated with an example. After the query-based primary retrieval, the \textit{analyzer} assesses the current task scenarios. Utilizing the analysis results, the \textit{organizer} dynamically provides execution plans and configurations, covering various potential workflows. Then the \textit{executor} carries out the plan and generate the final answer.}
    \label{fig:arch}
\end{figure*}

To address these limitations, we propose OkraLong, a novel retrieval-augmented framework that systematically optimizes long-text query answering. Unlike above approaches that rely on fixed workflow patterns, OkraLong flexibly orchestrates various pipelines according to different task scenarios.
As illustrated in Figure ~\ref{fig:arch}, our framework comprises three synergistic components: (1) Analyzer: a fine-tuned lightweight model that proactively characterizes task states, through query semantics and preliminarily retrieved contexts; (2) Adaptive Organizer: a dynamic scheduler that generates optimized execution plans, based on previous real-time analysis; (3) Compositive Executor: a modular operator suite that supports the execution of diverse processing pipelines and strategies.

Distinct from prior adaptive RAG methods (\citealp{flare}; \citealp{adarag}) that make coarse-grained decisions (e.g. whether to generate iteratively or retrieve additional data) , OkraLong is designed to fine-grainedly optimize the entire processing workflow, covering multiple modules and various pipelines. 
First, OkraLong constructs the flexibility to tailor strategies to specific requirements of different tasks, improving accuracy performance. For example, comparative query tasks demand separate entity retrieval, while bridging queries trigger stepwise reasoning chains.  Second, for financial effectiveness, our cost-aware scheduling dynamically allocates token budgets and information resources across various scenarios. For instance, semantically dense queries (e.g., abstractive synthesis) receive multiple aggregated contexts, whereas information-specific requests (e.g., fact extraction) utilize targeted context slicing. This allocation substantially reduce superfluous expenses. It is also worth noting that these flexible optimizations are facilitated by the task-understanding analyzer and we also develop several innovative execution operators to support tailored strategies. 

We evaluate OkraLong using an extensive collection of long-text understanding datasets, spanning multiple domains, ranging from simple factual queries to complex multi-step reasoning tasks.  The experimental results demonstrate that OkraLong not only enhances answer accuracy compared to existing advanced approaches, but also provides superior cost-effectiveness.

\section{Related Work}
\subsection{Long-Text Processing}
Understanding and reasoning over long-form text have always been crucial in natural language processing. 
Considerable efforts have been made to enhance LLMs to handle long contexts (\citealp{context_extend}; \citealp{unleashing}; \citealp{chen2023extending}).
Besides, increasingly powerful LLMs such as Gemini-1.5 \citep{team2024gemini} and GPT-4 \citep{achiam2023gpt}, have achieved remarkable large context capability, yet directly processing full-length content incurs high financial expenses. 

To address this issue, context compression has emerged as a practical solution for handling large prompts. Extractive compression methods directly select informational tokens or sentences from the context. For instance, RECOMP-Extr \citep{recomp} performs sentence-level selection based on similarity scores, while LLMlingua (\citealp{llmlingua}; \citealp{pan-etal-2024-llmlingua2}) and Longllmlingua \citep{longllmlingua} employ token-level filtering through information entropy. Abstractive approaches leverage generative models for content summarization, as exemplified by RECOMP-Abst\citep{recomp}, CompAct \citep{compact}, and Refiner \citep{refiner}. However, these methods still exhibit critical limitations : (1) heavily calling auxiliary models that introduce latency overhead, (2) potential loss of specific information during compression.

\subsection{Retrieval Augmented Generation}
Retrieval Augmented Generation (RAG) is a prevalent technique for enhancing LLM capabilities with external knowledge \citep{rag}.
Conventional RAG pipelines always segment voluminous text into chunks, embed them using lightweight models, and retrieve query-relevant content for the LLM generation \citep{gao2023rag-survey}.  
However, basic RAG systems are prone to information loss, particularly in multi-hop queries, leading to suboptimal accuracy (\citealp{tang2024multihoprag}; 
\citealp{zhuang-etal-2024-efficientrag}; \citealp{shao-etal-2023-iterative-rag}). 

Recent advances propose iterative and adaptive refinement mechanisms to mitigate these issues. For instance, FLARE \citep{flare}  and DRAGIN \citep{dragin} activate the search engine when LLMs output tokens with low probability. Self-RAG \citep{selfrag} and MIGRES \citep{migres} prompt LLMs to make decision on iterative retrieval. Adaptive-RAG \citep{adarag} and MBA-RAG \citep{mbarag} employ adaptive routing strategies to enhance effectiveness. \citet{self_ask} and \citet{self_correct} improve RAG performance utilizing self-asking and self-correcting. 
Despite these advancements, in practical settings, existing iterative methods often incur high costs due to extensive LLM calls, and the adaptive strategies remain coarse-grained, failing to account for diverse application scenarios.

\section{Methodology}
\subsection{Framework Overview}

In this section, we introduce OkraLong, a flexible and efficient retrieval-augmented framework for long-text query processing.
As depicted in Figure~\ref{fig:arch}, OkraLong comprises three core modules: analyzer, organizer and executor. 
Given a query and the long-form text, OkraLong initiates with primary query-relevant context retrieval. The context and the query are subsequently fed to our lightweight analyzer (implemented as a fine-tuned language model) for real-time task characterization. 
This analysis covers multiple dimensions, including task taxonomy, evidence containing and information patterns. These analytical outputs then drive our organizer to dynamically orchestrate specific execution plans with corresponding configurations. Finally, the executor conducts the optimized processing pipeline through a composition of operators, ultimately generating the final response.

With this architecture, OkraLong facilitates flexible and efficient processing workflows. For flexibility, we develop multiple execution operators as the core infrastructure, and the analyzer provides a comprehensive characterization of situational states. These allow for adaptive and fine-grained organization.
For efficiency, OkraLong orchestrates the appropriate processing mechanisms and budgets, optimizing both response quality and cost-effectiveness. Besides, the lightweight analyzer and compact scheduling pattern prevent unaffordable latency, an issue often neglected in previous iterative and generative methodologies.

\subsection{Analyzer}
The Analyzer constitutes the cognitive foundation of OkraLong, performing real-time assessment of the task states. Its analytical outputs drives dynamic workflow scheduling.

Initially, the analyzer activates the retriever to fetch query-relevant contexts, dispatched alongside the query for assessment. Compared to the query-centric approach, this inclusive strategy allows for a holistic assessment of both the query requirements and the overall task environment.

The derived analysis results cover three key aspects: (1) Task taxonomy: Queries with differing objectives demand diverse processing techniques. To organize these, we classify the tasks into five categories: arithmetic, extractive, abstractive, multi-source, and multi-bridge (multi-source tasks require information from various entities or sources, whereas multi-bridge tasks involve several interconnected procedural steps). (2) Information Pattern: The requisite information can manifest in different forms, such as semantically correlated or exactly matched patterns. We classify these patterns as either semantic, exact, or a combination of both, utilizing it to enhance retrieval strategies. (3) Evidence Identification: We confirm whether the initially retrieved context includes clear evidence to address the query, a step that also reveals the task's complexity.

Therefore, given long-from text $D$ and the query $q$, the analyzer can be formulated as:
\begin{equation}
C_q=\{c_1,c_2,...,c_k\}=\texttt{Retriever}(q,D)
\end{equation}
\begin{equation}
\{\theta_q, \psi_{i}, \phi_{e}\}= \texttt{Analyzer}(\texttt{Instruct}(q,C_q))
\end{equation}
where the query $q$ and the retrieved chunks $C_q$ are structured via the instructing prompt, and the $\theta_q, \psi_{i}, \phi_{e} $ represent the query, information and evidence terms in analytical results.

To implement the analyzer, we refine a light-weight language model via supervised fine-tuning. The training dataset is derived from public datasets. Entries with human annotations are processed and integrated, while unannotated entries are labeled utilizing the advanced LLM like GPT-4o (more details in Appendix~\ref{sec:train-details}).

\subsection{Executor}
The executor functions as the core operational engine of OkraLong, comprising multiple distinct operators. For clarity, we will discuss the execution component before delving into the organizer.

The primary role of the executor is to accomplish retrieval-augmented processing, which necessitates basic modules: indexing, retrieval, and generation. To facilitate more adaptive processing pipelines, we enhance these foundational elements and develop the following operators:
\begin{itemize}[leftmargin=1em,itemsep=1pt,topsep=5pt,parsep=2pt]
\item Fundamental Operators: Basic text chunking, indexing, context retrieval, and LLM generation.
\item Assembled Retriever: Catering to diverse information patterns, this operator integrates multiple retrieval strategies. It normalizes the matching scores and performs weighted aggregation to produce better context.
\item Context Processor: Instead of merely concatenating retrieved text chunks, this operator enables context merging, context extension, and table-recovery functionalities, ensuring both precise detail matching and enriched contextual information.
\item Query Splitter: For queries spanning multiple entities, this tool divides the main query into sub-queries, processed independently and subsequently aggregated to deliver a comprehensive response.
\item Step-wise Reasoner: Complex queries requiring reasoning benefit from this operator, which fosters step-by-step reasoning. Drawing inspiration from adaptive and iterative strategies, it prompts the LLM to execute reasoning steps, producing content for subsequent retrieval operations.
\end{itemize}

Through these operators, guided by the organizer's execution plans and configurations, the executor enables flexible pipeline construction. It supports both linear processing flows and complex branching topologies, adapting to diverse task characteristics.




\subsection{Organizer}
\label{sec:org}
The Organizer serves as the pivotal decision engine that transforms analytical insights into executable plans. It employs a task-aware heuristic orchestration to dynamically optimize and configure the processing pipeline. The organization covers three critical dimensions: workflow architecture, retrieval granularity, and evidence aggregation.

Workflow construction utilizes query-taxonomy ($\theta_q$) to organize task-specific processing pipelines, enhancing the targeted handling of queries. For example, multi-bridge queries, which require sequential reasoning across interdependent facts, are managed by a step-wise iterative pipeline that decomposes the task into chained sub-tasks with interleaved retrieval. Multi-source queries trigger a split-aggregate pipeline that independently processes evidence retrieval for distinct entities before final aggregation. Arithmetic queries activate an pipeline with context-extension after the specific retrieval, ensuring both precise detail matching and expansive contextual inclusion.  This typology-adaptive orchestration guarantees an efficient alignment between query complexity and processing strategy, enhancing both flexibility and efficiency.  

Retrieval granularity is governed through a dual-criteria mechanism adaptively. First, the characteristic of the query task ($\theta_q$) determines the primary retrieval scope and granularity  via a heuristic policy. Contextual tasks (e.g., abstractive queries) would activate extensive context scope, whereas factoid tasks (e.g., extractive) and iterative augmented tasks (e.g., multi-bridge) adopt more focused narrow spans. Additionally, the evidence state ($\phi_{e}$) triggers dynamic granularity adjustments: insufficient evidence initiates a scale extension and granularity expansion to incorporate broader evidential information.

Evidence aggregation integrates various retrieval strategies, applying tailored weights. Guided by the analyzed information pattern $\psi_{i}$, for tasks dominated by lexical patterns (e.g., entity lookup), the exact retrieval scores are boosted via weight $w_e$. Conversely, semantic-centric tasks see an upweighting in their semantic component (with $w_s$). 
The aggregated relevant score $S = w_e \cdot S_{exact} + w_s \cdot S_{semantic}$, produces an assembled retrieval output, adaptively adjusted to the information characteristics.

Overall, the organizer optimizes both the processing workflow and the 
 modular configurations. This strategic and flexible approach is fundamental to the framework's ability to robustly manage complex task scenarios.
\section{Experiment Setups}
\subsection{Datasets}
To evaluate the performance of  OkraLong comprehensively, we conduct experiments on six long-text query-answering datasets, spanning various domains and multiple query types: 

(1) \textbf{FINQA} \citep{chen2021finqa} is a finanicl numerical reasoning dataset, constructed from the public earnings reports of S\&P 500 companies. (2) \textbf{TAT-DQA} \citep{zhu2022tatdqa} is another finanical dataset, derived from annual reports. It incorporates a wide variety of query types such as fact extraction, counting, and arithmetic operations. (3) \textbf{Qasper} \citep{qasper} is a reading comprehension dataset based on NLP research papers, featuring both extractive and abstractive question types.
(4) \textbf{MultifieldQA} \citep{bai-etal-2024-longbench} has question-answering pairs sourced from diverse fields, including legal documents, government reports, encyclopedias, etc. (5) \textbf{HotpotQA} \citep{yang-etal-2018-hotpotqa} involves two-hop questions based on Wikipedia paragraphs.
(6) \textbf{2WikiMultihopQA} \citep{ho-etal-2020-2wiki} consists of up to five-hop questions, also based on Wikipedia content.

To align with practical long-text query settings such as unsegmented full content,
we utilize processed versions of the FINQA, TAT-DQA, and Qasper datasets from the UDA collection \citep{uda}. The MultifieldQA, HotpotQA, and 2WikiMultihopQA datasets are used in their corresponding forms from LongBench \citep{bai-etal-2024-longbench} (more details in Appendix ~\ref{sec:more-datasets}).

\subsection{Baselines and Setups}

\paragraph{\textit{Baselines.}} We select the following six approaches as the baselines: (1) \textbf{Standard RAG} utilizes a traditional chunking, retrieval, and generation workflow \citep{gao2023rag-survey}.
(2) \textbf{Long-Context Strategy} \citep{xu2024long_context} processes the entire long-text using a LLM without additional context refinement.
(3) \textbf{LongLLMLingua} \citep{longllmlingua}, a context compression approach that filters tokens based on informational significance according to a lightweight LLM.
(4) \textbf{CompAct} \citep{compact}, another compression approach, employing a lightweight LLM to iteratively generate the summarized text content.
(5) \textbf{FLARE} \citep{flare}, a dynamic RAG method that adapts retrieval based on token probabilities during iterative text generation.
(6) \textbf{Adaptive-RAG} \citep{adarag}, another dynamic RAG method, adaptively conducting multi-step or single-step retrieval based on query complexity.

\paragraph{\textit{Evaluation Metrics.}} To assess the quality of the  generated responses, we adopt the original evaluation metrics from the source benchmarks (\citealp{uda}; \citealp{bai-etal-2024-longbench}).
For FINQA's numerical-oriented tasks, we employ Exact Match (EM) accuracy, while using F1 scores for all other datasets. Additionally, we estimate the financial expenses by measuring the total token usage,  with output tokens assigned a four times higher cost-weight than input tokens \citep{openai-price}. Latency overhead is recorded as end-to-end execution time from query submission to final response generation.

\paragraph{\textit{Implementation Details.}} We utilize the GPT-4o \citep{hurst2024gpt} as the  backbone model for generating responses. Following the previous works (\citealp{recomp}; \citealp{selfrag}), we use the Contriever-MSMARCO \citep{izacard2021unsupervised} as the basic retrieval model, with the context chunk-size of 512 tokens. For the core-retrieval pipelines, the top-5 chunks are fetched, and in the compression-based pipelines, we fed the top-30 chunks to subsequent compression stages \citep{compact}. Within our OkraLong framework, we employ BM-25 \citep{robertson2009probabilistic} for exact retrieval augmentation.
And we perform supervised fine-tuning on the Llama-3.2-1B-Instruct \citep{dubey2024llama} model to serve as the lightweight analyzer. The fine-tuning dataset combines sampled train-splits from the HotpotQA, TAT-DQA, and Qasper datasets.
For more implementation details, please refer to Appendix~\ref{more implementation}.


\section{Results and Analysis}

\subsection{Main Results}

\begin{table*}[t]
\centering
\resizebox{\linewidth}{!}{
\setlength{\tabcolsep}{4pt}
\begin{tabular}{lcccccccccccccc}
\toprule
\multirow{2}{*}{Method} & \multicolumn{2}{c}{\textbf{Average}} & \multicolumn{2}{c}{TAT-DQA} & \multicolumn{2}{c}{FINQA} & \multicolumn{2}{c}{Qasper} & \multicolumn{2}{c}{M-FieldQA} & \multicolumn{2}{c}{HotpotQA} & \multicolumn{2}{c}{2WikiMQA} \\
\cmidrule(){2-3} \cmidrule(lr){4-5} \cmidrule(lr){6-7} \cmidrule(lr){8-9} \cmidrule(lr){10-11} \cmidrule(lr){12-13} \cmidrule(lr){14-15}
& 
\multicolumn{1}{c}{\textbf{Score}} & \multicolumn{1}{c}{\textbf{Cost}} & \multicolumn{1}{c}{F1} & \multicolumn{1}{c}{Cost} & \multicolumn{1}{c}{EM} & \multicolumn{1}{c}{Cost} & \multicolumn{1}{c}{F1} & \multicolumn{1}{c}{Cost} & \multicolumn{1}{c}{F1} & \multicolumn{1}{c}{Cost} & \multicolumn{1}{c}{F1} & \multicolumn{1}{c}{Cost} & \multicolumn{1}{c}{F1} & \multicolumn{1}{c}{Cost} \\
\midrule
Std-RAG & 46.2 & 2.5 & 43.3 & 2.9 & 45.0 & 2.8 & 33.8 & 2.4 & \underline{55.6} & 2.2 & 47.0 & 2.2 & 52.3 & 2.3 \\
LC & \textbf{57.8} & \textcolor{gray}{32.0} & \textbf{54.4} & 79.8 & \textbf{56.5} & 74.1 & \textbf{44.9} & 10.5 & \textbf{56.9} & 7.2 & \textbf{63.7} & 12.9 & \textbf{70.1} & 7.4 \\
\midrule
Longllmlingua & 42.1 & 2.6 & 43.5 & 3.5 & 41.4 & 3.4 & 34.5 & 2.5 & 39.5 & 1.9 & 54.0 & 2.4 & 39.7 & \underline{1.9} \\
Compact & \textcolor{gray}{30.9} & \textbf{0.5} & 19.0 & \textbf{0.6} & 11.9 & \textbf{0.6} & 19.4 & \textbf{0.3} & 44.3 & \textbf{0.3} & 45.8 & \textbf{0.5} & 45.0 & \textbf{0.5} \\
Ada-RAG & 48.6 & 3.9 & 42.6 & 5.3 & 43.9 & 4.9 & 36.5 & 2.7 & 51.2 & 2.4 & 56.3 & 3.9 & 60.9 & 4.2 \\
FLARE & 36.4 & 9.0 & 31.7 & 11.2 & 41.4 & 10.9 & 34.8 & 10.4 & 45.9 & 7.4 & 37.7 & 6.8 & 27.0 & 7.6 \\
\midrule
OkraLong  & \underline{51.4} & \underline{1.9} & \underline{53.3} & \underline{1.9} & \underline{45.3} & \underline{2.4} & \underline{36.6} & \underline{1.8} & 51.0 & \underline{1.5} & \underline{59.5} & \underline{1.9} & \underline{62.8} & 2.2 \\
\quad w/ Precise Mode  & \cellcolor{cyan!15}{\textbf{57.8}} & \cellcolor{cyan!15}{7.2} & 56.7 & 9.5 & 56.1 & 17.8 & 44.4 & 4.3 & 53.5 & 1.6 & 63.4 & 3.1 & 72.5 & 2.8 \\
\bottomrule
\end{tabular}
}
\caption{
End-to-end query answering performance across six datasets. Evaluation scores (F1/EM) are normalized to 0-100 scale for clarity, with the cost quantified as  token consumption ($\times10^{3}$ tokens) for LLM generation. Performance rankings are indicated with \textbf{bold} (for best) and \underline{underline} (for second best), where the augmented OkraLong with Precise-Mode is independently marked.
}
\label{tab:main_res}
\end{table*}

\begin{figure}[t]
  \includegraphics[width=\columnwidth]{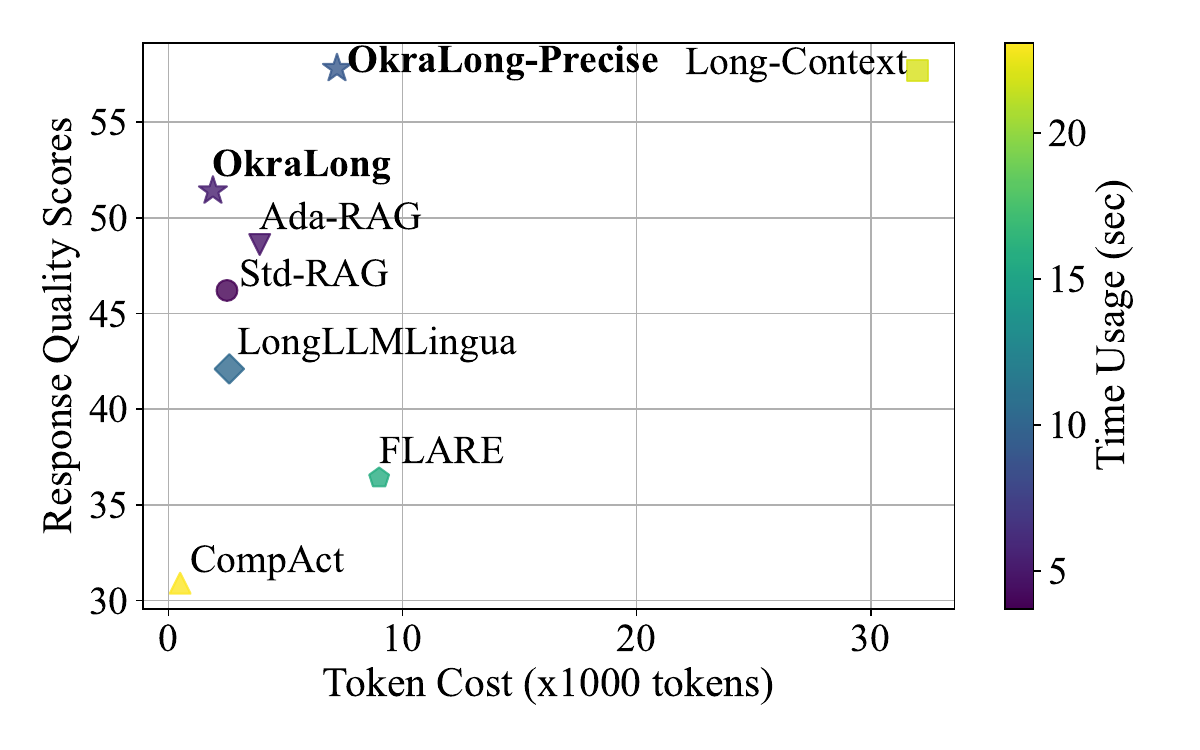}
  \caption{Average performance of end-to-end query answering across six datasets. Superior approaches are \textbf{left} and \textbf{top} positioned, indicating lower cost and higher accuracy. And the execution time is represented by the colors (the dark color denotes reduced latency).}
  \label{fig:res}
\end{figure}

The main results of our experimental evaluations are presented in Table~\ref{tab:main_res}, and a visualizing representation of the averaged results is displayed in Figure~\ref{fig:res}. Overall, OkraLong demonstrates significant effectiveness in both answering accuracy and cost efficiency.

Notably, the basic OkraLong maintains decent performance across diverse datasets, despite occasional suboptimal results on Multi-FieldQA. This may be attributed to the implicit semantic patterns among its factoid questions, which occasionally challenges the OkraLong's analyzer in task characterization.

Analysis of the baselines reveals two extreme cases: While compression-based CompAct minimizes token usage, its aggressive content summarization causes severe information loss (with 39.9\% accuracy degradation). Conversely, leveraging the 
capability of GPT-4o, long-context processing achieves peak accuracy through exhaustive context retention, but incurs prohibitive costs (16.8x higher than the basic OkraLong).

Aside from these extremes, the basic OkraLong enhances answer accuracy by \textbf{5.7\%-41.2\%} while achieving cost savings of \textbf{1.3x-4.7x} compared to prior advancement.
Furthermore, we also introduce the OkraLong with \texttt{Precise-Mode}, which automatically apply full context to initially unanswerable queries \footnote{The generating LLM is prompted to respond "unanswerable" if encountering a lack of evidence.}. This cascading augmentation achieves the equivalent answer quality with the long-context processing, while maintaining a \textbf{4.4x} cost advantage. The integration of both modes establish a \textbf{Pareto-optimal frontier} in the cost-accuracy spectrum (shown in Figure~\ref{fig:res}), enabling highly efficient deployment in practical long-text query processing.

\subsection{Latency Overhead}

\begin{figure}[t]
  \includegraphics[width=\columnwidth]{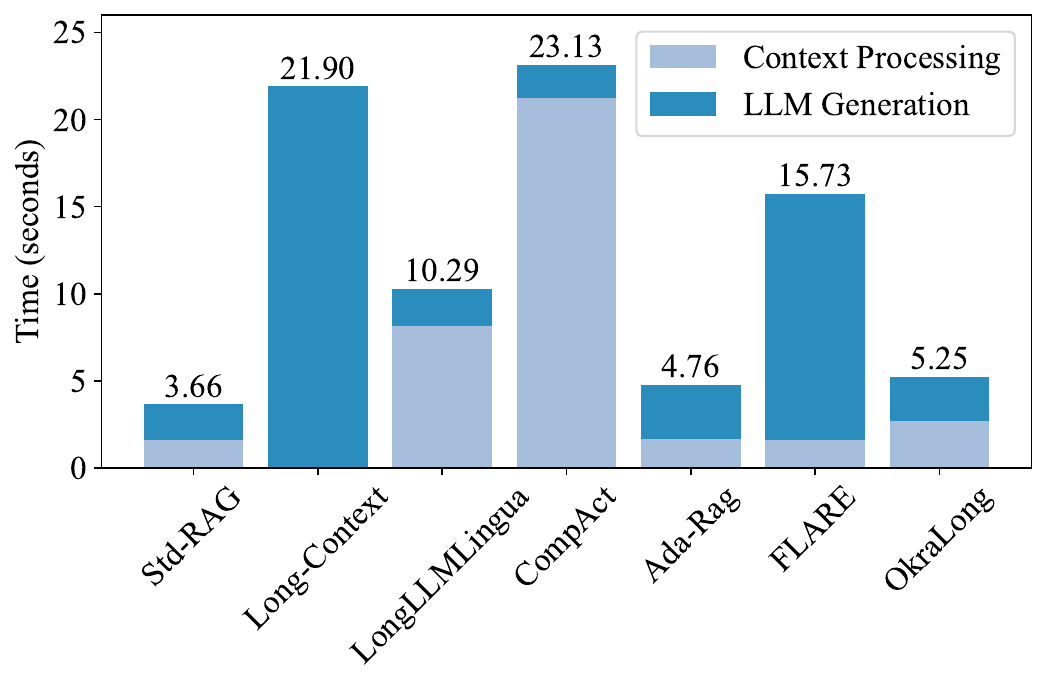}
  \caption{Average end-to-end latency results across various methods. The execution time (per query) comprises context processing time and LLM generation time. }
  \label{fig:lat}
\end{figure}

We conduct a comprehensive latency analysis among our OkraLong framework and 
other approaches. Figure~\ref{fig:lat} summarizes the average end-to-end latency results with decomposition.

The overall execution time can be divided into two primary components: context processing and LLM generation. For traditional approaches, the context processing with the standard RAG entails basic indexing and retrieval operations. The long-context mechanism requires no operations on context, but encounter substantial delays during LLM generation with lengthy input. For compression-based methods, the latency increases as they often rely on local models for iterative compression. Meanwhile, dynamic RAG approaches involve iterative LLM service calls, extending the generation time.Our OkraLong framework adaptively adjusts the workflow with a modest overhead of 0.19s for state analysis and 0.92s for re-scaled indexing. Given the improvements in accuracy and cost efficiency, this marginally extra time is justifiable.


More specific results across various datasets are shown in Table~\ref{tab:time_data}. The standard RAG method typically exhibits the lowest latency overhead, but its rigid processing leads to information loss.
The financial reports in TAT-DQA and FINQA datasets spanning hundreds of pages, 
which raises the latency overhead across all methods due to heavy indexing or lengthy full context. 
For HotpotQA and 2Wikimqa, which require multi-step reasoning, OkraLong spends more time than Qasper and M-FieldQA  due to iterative LLM calls. This also reflects the OkraLong's capacity to adapt to diverse demanding.

\begin{table}[t]
    \centering
    \resizebox{\linewidth}{!}{
    \begin{tabular}{lccccccc}
        \toprule
        Method & TAT & FIN & Qasper & M-Field & Hotpot & 2Wiki \\
        \midrule

        Std-RAG & \textbf{7.2} & \textbf{7.2} & \textbf{2.0} & \textbf{1.7} & \textbf{2.0} & \textbf{1.9} \\
        LC & 55.5 & 57.4 & 4.7 & 3.5 & 6.6 & 3.7 \\
        \midrule
        L-Lingua & 16.5 & 15.8 & 8.5 & 6.1 & 8.5 & 6.3 \\
        CompAct & 29.7 & 32.9 & 22.3 & 17.6 & 20.1 & 16.3 \\
        Ada-RAG & \underline{8.9} & \underline{7.7} & 3.2 & 2.4 & \underline{3.6} & \underline{2.8} \\
        FLARE & 31.3 & 27.7 & 14.6 & 6.9 & 6.0 & 7.8 \\
       \midrule
        OkraLong & \underline{8.9} & 10.5 & \underline{2.3} & \underline{2.3} & \underline{3.6} & 3.7 \\
        \bottomrule
    \end{tabular}
    }
    \caption{End-to-end execution latency of different methods across six datasets. Performance rankings are indicated with \textbf{bold} (for best) and \underline{underline} (for second best).}
    \label{tab:time_data}
\end{table}

\subsection{Ablation Study}
We conduct ablation studies to assess the contribution of various optimizations within OkraLong, using the TAT-DQA dataset for its extensive task characterization. Table~\ref{tab:ablation} displays the performance variations  when removing a specific optimization.

First, disabling adaptive workflow orchestration (i.e., using a fixed retrieval-generation pipeline for all tasks) reduces accuracy by 7.3\%. While this simplified version reduces token costs, it critically lacks situational adaptability, leading to inefficient processing for diverse query types.
Second, maintaining a fixed moderate retrieval granularity, without dynamically adjusting, results in a 14.1\% decrease in accuracy. This significant loss is due to the omission of critical information, especially for contextual tasks.
Third, replacing our aggregated retrieval approach with a direct dense retriever causes a 6.4\% F1 score drop.
This decline is primarily attributed to the limited capability of a single dense retriever to exactly fetch evidence for specific query details.

These results demonstrate that our synergistic approaches each provide different yet complementary benefits. The integration of multiple optimizations enhances the overall performance of OkraLong.

\begin{table}[t]
    \centering
    \resizebox{0.93\linewidth}{!}{
    \begin{tabular}{lcc}
        \toprule
        Method & F1 Score & Cost   \\
        \midrule
Std-RAG (baseline)   &   43.3    &   2.94   \\
\midrule
OkraLong & \textbf{53.3} & 1.93\\
\quad w/o workflow orchestration & 49.4 &  \textbf{1.59} \\
\quad w/o retrieval adjustment & 45.8 & 1.95  \\ 
\quad w/o aggregated retrieval & 49.9 & 1.91 \\
        \bottomrule
    \end{tabular}
    }
    \caption{Ablation studies on the OkraLong framework, assessing the contributions of diverse optimizations.}
    \label{tab:ablation}
\end{table}

\subsection{Analyzer Performance}
The analyzer serves as the cognitive foundation in OkraLong. In this section, we evaluate its performance across three terms: query type classification, information paradigm prediction, and evidence identification. Using the combined validation datasets, the prediction results of the analyzer is shown in Table~\ref{tab:analyzer}.

Compared to directly prompting the small model, supervised fine-tuning significantly improves the prediction performance. 
We observe that the query-type classification achieves a relatively high precision of 86.4\%, as a non-trivial  five-class categorization task. This efficiency aids in constructing the appropriate workflows, thereby enhancing the overall framework performance. The evidence identification also exhibits great performance with an exact-match score of 79.4\%. This facilitates the effective retrieval through dynamic scope adjustment and granularity control. However, the prediction on information paradigm shows reduced effectiveness. We attribute this to the inherent complexity to directly predict the optimal retrieval pattern (exact, semantic or both) through the query and contexts, which could exceed the capabilities of a lightweight language model. To mitigate potentially significant biases, we adopt conservative fusion-weights when integrating the two retrieval strategies.

\begin{table}[t]
    \centering
    \resizebox{\linewidth}{!}{
    \begin{tabular}{lccc}
        \toprule
         & Query & Information  & Evidence  \\
        \midrule
Analyzer    &   86.4    &    64.6      &    79.4 \\
 \quad w/o fine-tuning   &   22.9    &    15.2      &    67.5 \\
        \bottomrule
    \end{tabular}
    }
    \caption{Prediction accuracy (exact-match scores) across various terms, using fine-tuned analyzer or direct model answering. }
    \label{tab:analyzer}
\end{table}

\subsection{Robustness and Generalization}
In this section, we discuss the robustness and generalization of OkraLong across multiple dimensions. First, for the pattern robustness, the analyzer of OkraLong is fine-tuned on the heterogeneous dataset (combining HotpotQA, TAT-DQA and Qasper), mitigating over-fitting to specific patterns. The decent end-to-end Q\&A performance, across both in-distribution and three out-of-distribution unseen datasets, demonstrates the robustness in task understanding. Moreover, our evaluation comprehensively covers a wide range: (1) The target fields span finance, academia, government reports and general knowledge; (2) The long-text forms contain single document with varied lengths and concatenated multi-documents; (3) The query tasks ranging from factoid, arithmetic, to multi-step reasoning.  Leveraging OkraLong's synergistic components and optimizations, it exhibits effectiveness over this extensive range, showcasing its generalization across multiple scenarios.

Additionally, the modular design of the framework further enhances generalization. 
OkraLong allows new operations and heuristics to be flexibly integrated, accommodating diverse requirements and constraints.  For example, in scenarios where a small model with limited mathematical capabilities is employed, a code-aided generation operator can be seamlessly incorporated for arithmetic tasks. 
This adaptability ensures that OkraLong can be extended for diverse real-world application.

\section{Conclusion}
In this paper, we propose OkraLong, a flexible and efficient retrieval-augmented framework for long-text query processing. This innovative framework adaptively orchestrates the entire workflow through its three synergistic components: analyzer, organizer, and executor. OkraLong characterizes task states, dynamically configures the workflow, and carries out the execution to generate final answers.
 We conduct comprehensive evaluations across six diverse datasets, spanning multiple domains and complexity levels. The experimental results indicate that OkraLong not only enhances answering quality but also delivers significant cost-effectiveness.
Compared to pre-existing methods, OkraLong demonstrates superior performance in handling long-text queries, thereby providing a highly efficient solution for practical deployment.

\section*{Limitations}

While OkraLong demonstrates significant improvements in long-text query processing, we acknowledge its limitations: First, to balance accuracy and efficiency, the current analyzer employs supervised fine-tuning of a lightweight model,  which relies on annotated training datasets for refinement. Future research could explore semi-supervised or weakly supervised paradigms to further reduce annotation dependence while maintaining effectiveness. Second, while OkraLong efficiently processes textual content, some long-form documents may also require additional multi-modal integration. We currently focus on text-centric workflows, as it remains the primary information carrier. Exploring efficient strategies for querying long-form multi-modal content represents a promising direction for future work.
 
\bibliography{custom}

\begin{thebibliography}{45}
\providecommand{\natexlab}[1]{#1}

\bibitem[{Achiam et~al.(2023)Achiam, Adler, Agarwal, Ahmad, Akkaya, Aleman, Almeida, Altenschmidt, Altman, Anadkat et~al.}]{achiam2023gpt}
Josh Achiam, Steven Adler, Sandhini Agarwal, Lama Ahmad, Ilge Akkaya, Florencia~Leoni Aleman, Diogo Almeida, Janko Altenschmidt, Sam Altman, Shyamal Anadkat, et~al. 2023.
\newblock Gpt-4 technical report.
\newblock \emph{arXiv preprint arXiv:2303.08774}.

\bibitem[{Asai et~al.(2024)Asai, Wu, Wang, Sil, and Hajishirzi}]{selfrag}
Akari Asai, Zeqiu Wu, Yizhong Wang, Avirup Sil, and Hannaneh Hajishirzi. 2024.
\newblock \href {https://openreview.net/forum?id=hSyW5go0v8} {Self-{RAG}: Learning to retrieve, generate, and critique through self-reflection}.
\newblock In \emph{The Twelfth International Conference on Learning Representations}.

\bibitem[{AzureOpenAI(2025)}]{azure}
AzureOpenAI. 2025.
\newblock \href {https://learn.microsoft.com/en-us/azure/ai-services/openai/} {Azure openai service}.

\bibitem[{Bai et~al.(2024)Bai, Lv, Zhang, Lyu, Tang, Huang, Du, Liu, Zeng, Hou, Dong, Tang, and Li}]{bai-etal-2024-longbench}
Yushi Bai, Xin Lv, Jiajie Zhang, Hongchang Lyu, Jiankai Tang, Zhidian Huang, Zhengxiao Du, Xiao Liu, Aohan Zeng, Lei Hou, Yuxiao Dong, Jie Tang, and Juanzi Li. 2024.
\newblock \href {https://doi.org/10.18653/v1/2024.acl-long.172} {{L}ong{B}ench: A bilingual, multitask benchmark for long context understanding}.
\newblock In \emph{Proceedings of the 62nd Annual Meeting of the Association for Computational Linguistics (Volume 1: Long Papers)}, pages 3119--3137, Bangkok, Thailand. Association for Computational Linguistics.

\bibitem[{Chen et~al.(2023)Chen, Wong, Chen, and Tian}]{chen2023extending}
Shouyuan Chen, Sherman Wong, Liangjian Chen, and Yuandong Tian. 2023.
\newblock Extending context window of large language models via positional interpolation.
\newblock \emph{arXiv preprint arXiv:2306.15595}.

\bibitem[{Chen et~al.(2021)Chen, Chen, Smiley, Shah, Borova, Langdon, Moussa, Beane, Huang, Routledge et~al.}]{chen2021finqa}
Zhiyu Chen, Wenhu Chen, Charese Smiley, Sameena Shah, Iana Borova, Dylan Langdon, Reema Moussa, Matt Beane, Ting-Hao Huang, Bryan Routledge, et~al. 2021.
\newblock Finqa: A dataset of numerical reasoning over financial data.
\newblock \emph{arXiv preprint arXiv:2109.00122}.

\bibitem[{Chroma(2025)}]{chroma}
Chroma. 2025.
\newblock \href {https://github.com/chroma-core/chroma} {Chroma: the ai-native open-source embedding database}.

\bibitem[{Dasigi et~al.(2021)Dasigi, Lo, Beltagy, Cohan, Smith, and Gardner}]{qasper}
Pradeep Dasigi, Kyle Lo, Iz~Beltagy, Arman Cohan, Noah~A Smith, and Matt Gardner. 2021.
\newblock A dataset of information-seeking questions and answers anchored in research papers.
\newblock \emph{arXiv preprint arXiv:2105.03011}.

\bibitem[{Dubey et~al.(2024)Dubey, Jauhri, Pandey, Kadian, Al-Dahle, Letman, Mathur, Schelten, Yang, Fan et~al.}]{dubey2024llama}
Abhimanyu Dubey, Abhinav Jauhri, Abhinav Pandey, Abhishek Kadian, Ahmad Al-Dahle, Aiesha Letman, Akhil Mathur, Alan Schelten, Amy Yang, Angela Fan, et~al. 2024.
\newblock The llama 3 herd of models.
\newblock \emph{arXiv preprint arXiv:2407.21783}.

\bibitem[{Fei et~al.(2024)Fei, Niu, Zhou, Hou, Bai, Deng, and Han}]{fei-etal-2024-extend-lc}
Weizhi Fei, Xueyan Niu, Pingyi Zhou, Lu~Hou, Bo~Bai, Lei Deng, and Wei Han. 2024.
\newblock \href {https://doi.org/10.18653/v1/2024.findings-acl.306} {Extending context window of large language models via semantic compression}.
\newblock In \emph{Findings of the Association for Computational Linguistics: ACL 2024}, pages 5169--5181, Bangkok, Thailand. Association for Computational Linguistics.

\bibitem[{Gao et~al.(2024)Gao, Zhu, Cao, Zhou, Wu, Chen, Wu, Hu, and Dai}]{self_correct}
Yuan Gao, Yiheng Zhu, Yuanbin Cao, Yinzhi Zhou, Zhen Wu, Yujie Chen, Shenglan Wu, Haoyuan Hu, and Xinyu Dai. 2024.
\newblock \href {https://aclanthology.org/2024.lrec-main.476/} {Dr3: Ask large language models not to give off-topic answers in open domain multi-hop question answering}.
\newblock In \emph{Proceedings of the 2024 Joint International Conference on Computational Linguistics, Language Resources and Evaluation (LREC-COLING 2024)}, pages 5350--5364, Torino, Italia. ELRA and ICCL.

\bibitem[{Gao et~al.(2023)Gao, Xiong, Gao, Jia, Pan, Bi, Dai, Sun, and Wang}]{gao2023rag-survey}
Yunfan Gao, Yun Xiong, Xinyu Gao, Kangxiang Jia, Jinliu Pan, Yuxi Bi, Yi~Dai, Jiawei Sun, and Haofen Wang. 2023.
\newblock Retrieval-augmented generation for large language models: A survey.
\newblock \emph{arXiv preprint arXiv:2312.10997}.

\bibitem[{Ho et~al.(2020)Ho, Duong~Nguyen, Sugawara, and Aizawa}]{ho-etal-2020-2wiki}
Xanh Ho, Anh-Khoa Duong~Nguyen, Saku Sugawara, and Akiko Aizawa. 2020.
\newblock \href {https://doi.org/10.18653/v1/2020.coling-main.580} {Constructing a multi-hop {QA} dataset for comprehensive evaluation of reasoning steps}.
\newblock In \emph{Proceedings of the 28th International Conference on Computational Linguistics}, pages 6609--6625, Barcelona, Spain (Online). International Committee on Computational Linguistics.

\bibitem[{Hui et~al.(2024)Hui, Lu, and Zhang}]{uda}
Yulong Hui, Yao Lu, and Huanchen Zhang. 2024.
\newblock \href {https://openreview.net/forum?id=MS4oxVfBHn} {{UDA}: A benchmark suite for retrieval augmented generation in real-world document analysis}.
\newblock In \emph{The Thirty-eight Conference on Neural Information Processing Systems Datasets and Benchmarks Track}.

\bibitem[{Hurst et~al.(2024)Hurst, Lerer, Goucher, Perelman, Ramesh, Clark, Ostrow, Welihinda, Hayes, Radford et~al.}]{hurst2024gpt}
Aaron Hurst, Adam Lerer, Adam~P Goucher, Adam Perelman, Aditya Ramesh, Aidan Clark, AJ~Ostrow, Akila Welihinda, Alan Hayes, Alec Radford, et~al. 2024.
\newblock Gpt-4o system card.
\newblock \emph{arXiv preprint arXiv:2410.21276}.

\bibitem[{Hwang et~al.(2024)Hwang, Cho, Jeong, Song, Han, and Park}]{exit}
Taeho Hwang, Sukmin Cho, Soyeong Jeong, Hoyun Song, SeungYoon Han, and Jong~C Park. 2024.
\newblock Exit: Context-aware extractive compression for enhancing retrieval-augmented generation.
\newblock \emph{arXiv preprint arXiv:2412.12559}.

\bibitem[{Izacard et~al.(2021)Izacard, Caron, Hosseini, Riedel, Bojanowski, Joulin, and Grave}]{izacard2021unsupervised}
Gautier Izacard, Mathilde Caron, Lucas Hosseini, Sebastian Riedel, Piotr Bojanowski, Armand Joulin, and Edouard Grave. 2021.
\newblock Unsupervised dense information retrieval with contrastive learning.
\newblock \emph{arXiv preprint arXiv:2112.09118}.

\bibitem[{Jeong et~al.(2024)Jeong, Baek, Cho, Hwang, and Park}]{adarag}
Soyeong Jeong, Jinheon Baek, Sukmin Cho, Sung~Ju Hwang, and Jong Park. 2024.
\newblock \href {https://doi.org/10.18653/v1/2024.naacl-long.389} {Adaptive-{RAG}: Learning to adapt retrieval-augmented large language models through question complexity}.
\newblock In \emph{Proceedings of the 2024 Conference of the North American Chapter of the Association for Computational Linguistics: Human Language Technologies (Volume 1: Long Papers)}, pages 7036--7050, Mexico City, Mexico. Association for Computational Linguistics.

\bibitem[{Jiang et~al.(2023{\natexlab{a}})Jiang, Wu, Lin, Yang, and Qiu}]{llmlingua}
Huiqiang Jiang, Qianhui Wu, Chin-Yew Lin, Yuqing Yang, and Lili Qiu. 2023{\natexlab{a}}.
\newblock \href {https://openreview.net/forum?id=ADsEdyI32n} {{LLML}ingua: Compressing prompts for accelerated inference of large language models}.
\newblock In \emph{The 2023 Conference on Empirical Methods in Natural Language Processing}.

\bibitem[{Jiang et~al.(2024)Jiang, Wu, Luo, Li, Lin, Yang, and Qiu}]{longllmlingua}
Huiqiang Jiang, Qianhui Wu, Xufang Luo, Dongsheng Li, Chin-Yew Lin, Yuqing Yang, and Lili Qiu. 2024.
\newblock \href {https://doi.org/10.18653/v1/2024.acl-long.91} {{L}ong{LLML}ingua: Accelerating and enhancing {LLM}s in long context scenarios via prompt compression}.
\newblock In \emph{Proceedings of the 62nd Annual Meeting of the Association for Computational Linguistics (Volume 1: Long Papers)}, pages 1658--1677, Bangkok, Thailand. Association for Computational Linguistics.

\bibitem[{Jiang et~al.(2023{\natexlab{b}})Jiang, Xu, Gao, Sun, Liu, Dwivedi-Yu, Yang, Callan, and Neubig}]{flare}
Zhengbao Jiang, Frank Xu, Luyu Gao, Zhiqing Sun, Qian Liu, Jane Dwivedi-Yu, Yiming Yang, Jamie Callan, and Graham Neubig. 2023{\natexlab{b}}.
\newblock \href {https://doi.org/10.18653/v1/2023.emnlp-main.495} {Active retrieval augmented generation}.
\newblock In \emph{Proceedings of the 2023 Conference on Empirical Methods in Natural Language Processing}, pages 7969--7992, Singapore. Association for Computational Linguistics.

\bibitem[{Lewis et~al.(2020)Lewis, Perez, Piktus, Petroni, Karpukhin, Goyal, K\"{u}ttler, Lewis, Yih, Rockt\"{a}schel, Riedel, and Kiela}]{rag}
Patrick Lewis, Ethan Perez, Aleksandra Piktus, Fabio Petroni, Vladimir Karpukhin, Naman Goyal, Heinrich K\"{u}ttler, Mike Lewis, Wen-tau Yih, Tim Rockt\"{a}schel, Sebastian Riedel, and Douwe Kiela. 2020.
\newblock \href {https://proceedings.neurips.cc/paper_files/paper/2020/file/6b493230205f780e1bc26945df7481e5-Paper.pdf} {Retrieval-augmented generation for knowledge-intensive nlp tasks}.
\newblock In \emph{Advances in Neural Information Processing Systems}, volume~33, pages 9459--9474. Curran Associates, Inc.

\bibitem[{Li et~al.(2024{\natexlab{a}})Li, Cao, Ma, and Sun}]{li2024long}
Xinze Li, Yixin Cao, Yubo Ma, and Aixin Sun. 2024{\natexlab{a}}.
\newblock Long context vs. rag for llms: An evaluation and revisits.
\newblock \emph{arXiv preprint arXiv:2501.01880}.

\bibitem[{Li et~al.(2024{\natexlab{b}})Li, Hu, Liu, Zheng, Huang, and Xiong}]{refiner}
Zhonghao Li, Xuming Hu, Aiwei Liu, Kening Zheng, Sirui Huang, and Hui Xiong. 2024{\natexlab{b}}.
\newblock Refiner: Restructure retrieval content efficiently to advance question-answering capabilities.
\newblock \emph{arXiv preprint arXiv:2406.11357}.

\bibitem[{Li et~al.(2024{\natexlab{c}})Li, Li, Zhang, Mei, and Bendersky}]{li2024self-router}
Zhuowan Li, Cheng Li, Mingyang Zhang, Qiaozhu Mei, and Michael Bendersky. 2024{\natexlab{c}}.
\newblock Retrieval augmented generation or long-context llms? a comprehensive study and hybrid approach.
\newblock In \emph{Proceedings of the 2024 Conference on Empirical Methods in Natural Language Processing: Industry Track}, pages 881--893.

\bibitem[{Liang et~al.(2023)Liang, Wang, Huang, Wu, Wu, Lu, Ma, and Li}]{unleashing}
Xinnian Liang, Bing Wang, Hui Huang, Shuangzhi Wu, Peihao Wu, Lu~Lu, Zejun Ma, and Zhoujun Li. 2023.
\newblock Unleashing infinite-length input capacity for large-scale language models with self-controlled memory system.
\newblock \emph{arXiv e-prints}, pages arXiv--2304.

\bibitem[{Ma et~al.(2023)Ma, Gong, He, Zhao, and Duan}]{ma-etal-2023-rewrite}
Xinbei Ma, Yeyun Gong, Pengcheng He, Hai Zhao, and Nan Duan. 2023.
\newblock \href {https://doi.org/10.18653/v1/2023.emnlp-main.322} {Query rewriting in retrieval-augmented large language models}.
\newblock In \emph{Proceedings of the 2023 Conference on Empirical Methods in Natural Language Processing}, pages 5303--5315, Singapore. Association for Computational Linguistics.

\bibitem[{OpenAI(2025)}]{openai-price}
OpenAI. 2025.
\newblock \href {https://openai.com/api/pricing/} {Openai api pricing}.

\bibitem[{Pan et~al.(2024)Pan, Wu, Jiang, Xia, Luo, Zhang, Lin, R{\"u}hle, Yang, Lin, Zhao, Qiu, and Zhang}]{pan-etal-2024-llmlingua2}
Zhuoshi Pan, Qianhui Wu, Huiqiang Jiang, Menglin Xia, Xufang Luo, Jue Zhang, Qingwei Lin, Victor R{\"u}hle, Yuqing Yang, Chin-Yew Lin, H.~Vicky Zhao, Lili Qiu, and Dongmei Zhang. 2024.
\newblock \href {https://doi.org/10.18653/v1/2024.findings-acl.57} {{LLML}ingua-2: Data distillation for efficient and faithful task-agnostic prompt compression}.
\newblock In \emph{Findings of the Association for Computational Linguistics: ACL 2024}, pages 963--981, Bangkok, Thailand. Association for Computational Linguistics.

\bibitem[{Press et~al.(2023)Press, Zhang, Min, Schmidt, Smith, and Lewis}]{self_ask}
Ofir Press, Muru Zhang, Sewon Min, Ludwig Schmidt, Noah Smith, and Mike Lewis. 2023.
\newblock \href {https://doi.org/10.18653/v1/2023.findings-emnlp.378} {Measuring and narrowing the compositionality gap in language models}.
\newblock In \emph{Findings of the Association for Computational Linguistics: EMNLP 2023}, pages 5687--5711, Singapore. Association for Computational Linguistics.

\bibitem[{Robertson et~al.(2009)Robertson, Zaragoza et~al.}]{robertson2009probabilistic}
Stephen Robertson, Hugo Zaragoza, et~al. 2009.
\newblock The probabilistic relevance framework: Bm25 and beyond.
\newblock \emph{Foundations and Trends{\textregistered} in Information Retrieval}, 3(4):333--389.

\bibitem[{Shao et~al.(2023)Shao, Gong, Shen, Huang, Duan, and Chen}]{shao-etal-2023-iterative-rag}
Zhihong Shao, Yeyun Gong, Yelong Shen, Minlie Huang, Nan Duan, and Weizhu Chen. 2023.
\newblock \href {https://doi.org/10.18653/v1/2023.findings-emnlp.620} {Enhancing retrieval-augmented large language models with iterative retrieval-generation synergy}.
\newblock In \emph{Findings of the Association for Computational Linguistics: EMNLP 2023}, pages 9248--9274, Singapore. Association for Computational Linguistics.

\bibitem[{Su et~al.(2024)Su, Tang, Ai, Wu, and Liu}]{dragin}
Weihang Su, Yichen Tang, Qingyao Ai, Zhijing Wu, and Yiqun Liu. 2024.
\newblock \href {https://doi.org/10.18653/v1/2024.acl-long.702} {{DRAGIN}: Dynamic retrieval augmented generation based on the real-time information needs of large language models}.
\newblock In \emph{Proceedings of the 62nd Annual Meeting of the Association for Computational Linguistics (Volume 1: Long Papers)}, pages 12991--13013, Bangkok, Thailand. Association for Computational Linguistics.

\bibitem[{Tang et~al.(2025)Tang, Gao, Li, Du, Li, and Xie}]{mbarag}
Xiaqiang Tang, Qiang Gao, Jian Li, Nan Du, Qi~Li, and Sihong Xie. 2025.
\newblock \href {https://aclanthology.org/2025.coling-main.218/} {{MBA}-{RAG}: a bandit approach for adaptive retrieval-augmented generation through question complexity}.
\newblock In \emph{Proceedings of the 31st International Conference on Computational Linguistics}, pages 3248--3254, Abu Dhabi, UAE. Association for Computational Linguistics.

\bibitem[{Tang and Yang(2024)}]{tang2024multihoprag}
Yixuan Tang and Yi~Yang. 2024.
\newblock \href {https://openreview.net/forum?id=t4eB3zYWBK} {Multihop-{RAG}: Benchmarking retrieval-augmented generation for multi-hop queries}.
\newblock In \emph{First Conference on Language Modeling}.

\bibitem[{Team et~al.(2024)Team, Georgiev, Lei, Burnell, Bai, Gulati, Tanzer, Vincent, Pan, Wang et~al.}]{team2024gemini}
Gemini Team, Petko Georgiev, Ving~Ian Lei, Ryan Burnell, Libin Bai, Anmol Gulati, Garrett Tanzer, Damien Vincent, Zhufeng Pan, Shibo Wang, et~al. 2024.
\newblock Gemini 1.5: Unlocking multimodal understanding across millions of tokens of context.
\newblock \emph{arXiv preprint arXiv:2403.05530}.

\bibitem[{Trivedi et~al.(2023)Trivedi, Balasubramanian, Khot, and Sabharwal}]{trivedi-etal-2023-ircot}
Harsh Trivedi, Niranjan Balasubramanian, Tushar Khot, and Ashish Sabharwal. 2023.
\newblock \href {https://doi.org/10.18653/v1/2023.acl-long.557} {Interleaving retrieval with chain-of-thought reasoning for knowledge-intensive multi-step questions}.
\newblock In \emph{Proceedings of the 61st Annual Meeting of the Association for Computational Linguistics (Volume 1: Long Papers)}, pages 10014--10037, Toronto, Canada. Association for Computational Linguistics.

\bibitem[{Tworkowski et~al.(2023)Tworkowski, Staniszewski, Pacek, Wu, Michalewski, and Mi\l{}o\'{s}}]{context_extend}
Szymon Tworkowski, Konrad Staniszewski, Miko\l{}aj Pacek, Yuhuai Wu, Henryk Michalewski, and Piotr Mi\l{}o\'{s}. 2023.
\newblock Focused transformer: contrastive training for context scaling.
\newblock In \emph{Proceedings of the 37th International Conference on Neural Information Processing Systems}, NIPS '23, Red Hook, NY, USA. Curran Associates Inc.

\bibitem[{Wang et~al.(2024)Wang, Duan, Li, Wang, and Cai}]{migres}
Keheng Wang, Feiyu Duan, Peiguang Li, Sirui Wang, and Xunliang Cai. 2024.
\newblock Llms know what they need: Leveraging a missing information guided framework to empower retrieval-augmented generation.
\newblock \emph{arXiv preprint arXiv:2404.14043}.

\bibitem[{Xu et~al.(2024{\natexlab{a}})Xu, Shi, and Choi}]{recomp}
Fangyuan Xu, Weijia Shi, and Eunsol Choi. 2024{\natexlab{a}}.
\newblock \href {https://openreview.net/forum?id=mlJLVigNHp} {{RECOMP}: Improving retrieval-augmented {LM}s with context compression and selective augmentation}.
\newblock In \emph{The Twelfth International Conference on Learning Representations}.

\bibitem[{Xu et~al.(2024{\natexlab{b}})Xu, Ping, Wu, McAfee, Zhu, Liu, Subramanian, Bakhturina, Shoeybi, and Catanzaro}]{xu2024long_context}
Peng Xu, Wei Ping, Xianchao Wu, Lawrence McAfee, Chen Zhu, Zihan Liu, Sandeep Subramanian, Evelina Bakhturina, Mohammad Shoeybi, and Bryan Catanzaro. 2024{\natexlab{b}}.
\newblock \href {https://openreview.net/forum?id=xw5nxFWMlo} {Retrieval meets long context large language models}.
\newblock In \emph{The Twelfth International Conference on Learning Representations}.

\bibitem[{Yang et~al.(2018)Yang, Qi, Zhang, Bengio, Cohen, Salakhutdinov, and Manning}]{yang-etal-2018-hotpotqa}
Zhilin Yang, Peng Qi, Saizheng Zhang, Yoshua Bengio, William Cohen, Ruslan Salakhutdinov, and Christopher~D. Manning. 2018.
\newblock \href {https://doi.org/10.18653/v1/D18-1259} {{H}otpot{QA}: A dataset for diverse, explainable multi-hop question answering}.
\newblock In \emph{Proceedings of the 2018 Conference on Empirical Methods in Natural Language Processing}, pages 2369--2380, Brussels, Belgium. Association for Computational Linguistics.

\bibitem[{Yoon et~al.(2024)Yoon, Lee, Hwang, Jeong, and Kang}]{compact}
Chanwoong Yoon, Taewhoo Lee, Hyeon Hwang, Minbyul Jeong, and Jaewoo Kang. 2024.
\newblock \href {https://doi.org/10.18653/v1/2024.emnlp-main.1194} {{C}omp{A}ct: Compressing retrieved documents actively for question answering}.
\newblock In \emph{Proceedings of the 2024 Conference on Empirical Methods in Natural Language Processing}, pages 21424--21439, Miami, Florida, USA. Association for Computational Linguistics.

\bibitem[{Zhu et~al.(2022)Zhu, Lei, Feng, Wang, Zhang, and Chua}]{zhu2022tatdqa}
Fengbin Zhu, Wenqiang Lei, Fuli Feng, Chao Wang, Haozhou Zhang, and Tat-Seng Chua. 2022.
\newblock Towards complex document understanding by discrete reasoning.
\newblock In \emph{Proceedings of the 30th ACM International Conference on Multimedia}, pages 4857--4866.

\bibitem[{Zhuang et~al.(2024)Zhuang, Zhang, Cheng, Yang, Liu, Huang, Lin, Rajmohan, Zhang, and Zhang}]{zhuang-etal-2024-efficientrag}
Ziyuan Zhuang, Zhiyang Zhang, Sitao Cheng, Fangkai Yang, Jia Liu, Shujian Huang, Qingwei Lin, Saravan Rajmohan, Dongmei Zhang, and Qi~Zhang. 2024.
\newblock \href {https://doi.org/10.18653/v1/2024.emnlp-main.199} {{E}fficient{RAG}: Efficient retriever for multi-hop question answering}.
\newblock In \emph{Proceedings of the 2024 Conference on Empirical Methods in Natural Language Processing}, pages 3392--3411, Miami, Florida, USA. Association for Computational Linguistics.

\end{thebibliography}

\newpage
\newpage
\appendix
\section{Training Details of the Analyzer}

\label{sec:train-details}
\subsection{Dataset Construction.}
The training dataset is constructed from the sampled training splits of the  HotpotQA, TAT-DQA, and Qasper datasets. Detailed statistics are provided in Table~\ref{tab:train-stat}.

The input for each training instance includes the query and the related document segments, with a specific prompt outlined in Table~\ref{tab:ana_prompt}. 

Regarding the output terms: (1) Question types, annotated in the original datasets, are standardized into five predefined categories. (2) The information paradigm guides retrieval policy selection. We conduct retrieval with two independent strategies. The context containing more pertinent evidence is designated as the preferred paradigm (either exact, semantic, or same). (3) The evidence containing is directly labeled according to the annotated evidence and the provided contexts. For datasets lacking usable evidence annotations, such as TAT-DQA, evidence identification are labeled using GPT-4o.

\begin{table}[h]
    \centering
    \resizebox{0.6 \linewidth}{!}{
    \begin{tabular}{lcc}
        \toprule
          Dataset & \#Train  & \#Valid   \\
        \midrule
    TAT-DQA    &   3.2k     &    0.3k           \\
    HotpotQA  &  3.6k       & 0.4k \\ 
    Qasper &   2.6k  & 0.3k  \\
    Total &  9.4k & 1.0k  \\
        \bottomrule
    \end{tabular}
    }
    \caption{Statistics of the training dataset. The aggregated dataset size is restricted due to limited records available in Qasper and to maintain dataset balance.}
    \label{tab:train-stat}
\end{table}

 \subsection{Training Configuration}
 We conduct the supervised fine-tuning on Llama-3.2-1B-Instruct model, with LoRA using the following settings: 
 \begin{itemize}[leftmargin=2em,itemsep=2pt,topsep=5pt,parsep=2pt]
     \item Gradient accumulation steps: 64
     \item Learning rate: 1e-4
     \item Training epochs: 5
     \item LoRA rank: 8
     \item  LoRA scaling: 16
     \item  LoRA dropout: 0.1
 \end{itemize}

\section{More Implementation Details}
\subsection{Experimental Dataset}
\label{sec:more-datasets}

In our experiments, we utilize the datasets originating from the long-form aligned UDA collection \citep{uda} and LongBench collection \citep{bai-etal-2024-longbench}, adhering to their established configurations. 


UDA preserves the complete, unsegmented documents along with the source question-answering data points. And LongBench aggregates multiple Wikipedia articles to furnish expansive long-form contexts. The statistics of the test datasets is detailed in Table \ref{tab:test} , illustrating the distribution across two benchmarks.

\begin{table}[t]
    \centering
    \resizebox{\linewidth}{!}{
    \begin{tabular}{lccccccc}
        \toprule
        \multirow{2}{*}{Dataset} & \multicolumn{3}{c}{UDA} & \multicolumn{3}{c}{Long Bench} \\
        \cmidrule(lr){2-4} \cmidrule(lr){5-7}
        & TAT & FIN & Qasper & M-Field & Hotpot & 2Wiki \\
        \midrule
        Test Size & 210 & 278 & 232 & 150 & 200 & 200 \\
        \bottomrule
    \end{tabular}
    }
    \caption{Distribution of test sizes across different datasets, following the settings of the original benchmark collections.}
    \label{tab:test}
\end{table}

\subsection{Experimental Details}
\label{more implementation}

In our experiments, we employ the GPT-4o model through the AzureOpenAI API \citep{azure}, with the version of 2024-08-06. Additional open-source models are sourced from Huggingface. For our retrieval processes, we utilize ChromaDB \citep{chroma} as the vector database. The fine-tuning of our analyzer is conducted on an NVIDIA A100 GPU for an hour, while local model deployments operate on an NVIDIA A10 GPU. The above setup mirrors the general scenario where average individuals deploy lightweight models on limited-capacity GPUs while accessing more powerful LLMs via remote high-end servers or APIs. 

When conducting the long-context processing, issues may raise where the complete context surpasses the 128k token limit of the GPT-4o context window. In such cases, we implement a fallback strategy that involves retrieving the top 200 most relevant text chunks, approximately aggregating to 100k tokens, using a dense retriever.

\subsection{Implementation Details of OkraLong}
The execution module of OkraLong integrates multiple operators: (1) In the assembled retriever, we deploy dual retrieval strategies—dense retriever and sparse retriever. We normalize the relevance scores of the top 20 text chunks using min-max normalization and aggregate them based on semantic or exact match preferences. These preferences influence the final scoring, applying a conservative weight preference of 3:2 or 1:1 to determine the top-ranked chunks. (2) The context processor maintains essential metadata such as positions and index numbers of text chunks. Utilizing this metadata, it merges neighboring chunks and extends their preceding and succeeding contexts if required. Additionally, we have developed a heuristic mechanism to detect and recover incomplete tables within the text, leveraging structural markers such as spacing and line breaks. (3) Inspired by 
previous works (\citealp{trivedi-etal-2023-ircot}; \citealp{flare}; \citealp{ma-etal-2023-rewrite}), we prompt the LLM to perform query splitting and step-wise reasoning. The detailed instructions are shown in Table~\ref{tab:query_split} and Table~\ref{tab:step-wise}.

During the analysis and organization, OkraLong primarily retrieves three segments, each comprising 150 tokens, to perform analysis. Subsequent to this analysis, contextual tasks such as abstractive queries span eight segments, whereas factoid tasks cover five segments. In instances where evidence is analyzed to be absent, the granularity of retrieval adjusts to 400 tokens for contextual tasks and 256 tokens for factoid tasks. Additionally, beyond the predefined settings, a relative score threshold is also utilized, set at 0.1 compared to the top-1 segment, to efficiently exclude highly irrelevant segments.

\begin{table*}[t]
    \centering
    \small
    \renewcommand{\arraystretch}{1.4} 
    \begin{tabular}{p{0.965\linewidth}} 
        \toprule
        \textbf{System:}
        \\
        Given a question and the document context, please answer three questions:\\
1. What type of question is being asked?  The types include: extractive, abstractive, arithmetic,  multi-bridge, and multi-source.  Extractive means the query is directly factoid; abstractive means the query needs large context and refinement; arithemtic means the query needs numerical calculation; multi-bridge means the answer requires multiple bridging steps to get the answer;multi-source means the answer requires information from multiple facts (e.g. comparison questions).\\
2. Is the key information of the question more exact or semantic (according to both the question and the context)? The answer should be "exact", "semantic" or "same".\\
3. Does the provided context contain the enough information to answer the question? The answer should be either  "yes" or "no".\\
The final answer should be in the format of a dictionary:\\ \{"question-type": "extractive", "info-type": "exact", "containing": "yes"\}. \\Please strictly follow the format and no explanation is needed.\\
        \midrule
        \textbf{User:} \\
        \#\#\# Context:  \{context\} \quad \#\#\# Question: \{question\} \quad \#\#\# Answer: \\
        \bottomrule
    \end{tabular}
    \caption{The instructed prompt for the task analyzer.}
    \label{tab:ana_prompt}
\end{table*}

\begin{table*}[t]
    \centering
    \small
    \renewcommand{\arraystretch}{1.4} 
    \begin{tabular}{p{0.965\linewidth}} 
        \toprule
        \textbf{System:}
        \\
        Given a question, and this question may need the information from multiple sources.\\ Please split this question into multiple sub-questions, each of which can be answered by a single source. The final answer should be several sub-questions separated by the line-breaker.\\
        \midrule 
        \textbf{Demonstration:}\\
        User: \\
        Which university has the larger campus, University of New Haven or University of West Florida?\\
        Assistant: 
        \\What is the campus size of University of New Haven?
        \\ What is the campus size of University of West Florida? \\        
        \midrule
        \textbf{User:} \\
\{question\} \\ 
        \bottomrule
    \end{tabular}
    \caption{The instructed prompt for the query splitting operator.}
    \label{tab:query_split}
\end{table*}

\begin{table*}[t]
    \centering
    \small
    \renewcommand{\arraystretch}{1.4} 
    \begin{tabular}{p{0.965\linewidth}} 
        \toprule
        \textbf{System:}
        \\
        Given a question, which may need multiple steps to get the final answer. Please first get the existing evidence for the question based on the given context, and then generate a next-step query to query additional information. If the question can already be totally answered, you should output '\#\#\# Answer: The answer is: <answer>' at the end. Otherwise, output 'None'. The answer should be based only on the context. """\\
        \midrule 
        \textbf{Demonstration:}\\
        User: \\
        \#\#\# Context: 100 Rifles is directed by Tom Gries and starring Jim Brown and Raquel Welch.\quad \#\#\# Question: 100 Rifles is a western film, starring an actress of what nationality?\\
        Assistant: \\
        \#\#\# Evidence: The main actress in 100 Rifles is Raquel Welch. \quad \#\#\# Next-Query: What is the nationality of Raquel Welch? \quad \#\#\# Answer: None\\      
        \midrule
        \textbf{User:} \\
        \#\#\# Context: \{context\} \quad \#\#\# Question:\{question\} \\
        \bottomrule
    \end{tabular}
    \caption{The instructed prompt for the step-wise reasoning operator.}
    \label{tab:step-wise}
\end{table*}

\end{document}